\documentclass[sort&compress, numafflabel]{elsarticle}

\usepackage[]{natbib}

\usepackage[breaklinks,hidelinks]{hyperref}
\usepackage{times}
\usepackage{latexsym}

\usepackage[margin=1in]{geometry}

\usepackage{subfiles}
\usepackage{multirow}
\usepackage{array}
\usepackage{hyphenat}
\usepackage{subcaption}
\usepackage{longtable}
\usepackage{graphicx}
\usepackage{adjustbox}
\usepackage{enumitem}
\usepackage{url}

\usepackage[color=yellow, textsize=small]{todonotes}

\usepackage{bm}
\usepackage{booktabs}
\usepackage{float}
\usepackage[english]{babel}
\usepackage{blindtext}
\usepackage{textcomp}
\usepackage{soul,color}

\makeatletter
\def\ps@pprintTitle{%
 \let\@oddhead\@empty
 \let\@evenhead\@empty
 \def\@oddfoot{}%
 \let\@evenfoot\@oddfoot}
\makeatother

\usepackage{titlesec}
\titleformat{\section}
      {\normalfont\bfseries}
      {\thesection}
      {0ex}
      {\MakeUppercase}

\titleformat{\subsection}
      {\normalfont\bfseries}
      {\thesection}
      {0ex}
      {}

\titleformat{\subsubsection}
      {\normalfont}
      {\thesection}
      {0ex}
      {}

\usepackage{setspace}
\newif\ifsubfile
\subfiletrue

\newif\iftif
\tiffalse

\usepackage{microtype}

\usepackage{bm}

\usepackage{array}
\usepackage{amssymb}
\usepackage{pifont}
\newcommand{\cmark}{\ding{52}}%
\newcommand{\xmark}{\ding{54}}%

\newcommand{\challengeName}{n2c2/UW SDOH Challenge}

\newcommand\blfootnote[1]{%
  \begingroup
  \renewcommand\thefootnote{}\footnote{#1}%
  \addtocounter{footnote}{-1}%
  \endgroup
}

\title{The 2022 n2c2/UW Shared Task on Extracting Social Determinants of Health}

\author[add1]{Kevin Lybarger\textsuperscript{*}\blfootnote{\textsuperscript{*}Corresponding author: Kevin Lybarger, klybarge@gmu.edu.}}
\ead{}
\author[add2]{Meliha Yetisgen}
\ead{melihay@uw.edu}
\author[add1]{Özlem Uzuner}
\ead{}

\address[add1]{Department of Information Sciences and Technology, George Mason University, Fairfax, VA, USA 
}
\address[add2]{Department of Biomedical Informatics \& Medical Education, University of Washington, Seattle, WA, USA}

\begin{document}

\subfilefalse

\newpageafter{author}

\begin{abstract}

\noindent\textbf{Objective:} The \challengeName{} explores the extraction of social determinant of health (SDOH) information from clinical notes. The objectives include the advancement of natural language processing (NLP) information extraction techniques for SDOH and clinical information more broadly. This paper presents the shared task, data, participating teams, performance results, and considerations for future work. \\

\noindent\textbf{Materials and Methods:} The task used the Social History Annotated Corpus (SHAC), which consists of clinical text with detailed event-based annotations for SDOH events such as alcohol, drug, tobacco, employment, and living situation. Each SDOH event is characterized through attributes related to status, extent, and temporality. The task includes three subtasks related to information extraction (Subtask A), generalizability (Subtask B), and learning transfer (Subtask C). In addressing this task, participants utilized a range of techniques, including rules, knowledge bases, n-grams, word embeddings, and pretrained language models (LM). \\

\noindent\textbf{Results:} A total of 15 teams participated, and the top teams utilized pretrained deep learning LM. The top team across all subtasks used a sequence-to-sequence approach achieving 0.901 F1 for Subtask A, 0.774 F1 Subtask B, and 0.889 F1 for Subtask C. \\

\noindent\textbf{Conclusions:} Similar to many NLP tasks and domains, pretrained LM yielded the best performance, including generalizability and learning transfer. An error analysis indicates extraction performance varies by SDOH, with lower performance achieved for conditions, like substance use and homelessness, that increase health risks (risk factors) and higher performance achieved for conditions, like substance abstinence and living with family, that reduce health risks (protective factors). \\

\end{abstract}


\maketitle

\pagebreak


\section*{Background and Significance}

Social determinants of health (SDOH) are the conditions in which people live that affect quality-of-life and health, including social, economic, and behavioral factors.\cite{cdc_2021} SDOH include \textit{protective factors}, like social support, that reduce health risks and \textit{risk factors}, like housing instability, that increase health risks.\cite{alderwick2019meanings} SDOH are increasingly recognized for their impact on health outcomes and may contribute to decreased life expectancy.\cite{friedman2018toward, daniel2018addressing, himmelstein2018determined, singh2017social} As examples, substance abuse,\cite{centers2005annual, degenhardt2012extent, world2019global} living alone,\cite{cacioppo2003social, hawkley2015perceived}, housing instability,\cite{oppenheimer2016homelessness} and unemployment \cite{clougherty2010work, dooley1996health} are health risk factors. Knowledge of SDOH is important to clinical decision-making, improving health outcomes, and advancing health equity.\cite{singh2017social, blizinsky2018leveraging}

The Electronic Health Record (EHR) captures SDOH information through structured data and unstructured clinical notes. However, clinical notes contain more detailed and nuanced descriptions of many SDOH than are available through structured sources. Utilizing textual information from clinical notes in large-scale studies, clinical decision-support systems, and other secondary use applications, requires automatic extraction of key characteristics using natural language processing (NLP) information extraction techniques. Automatically extracted structured representations of SDOH can augment available structured SDOH data to produce more comprehensive patient representations.\cite{demner2009can, jensen2012mining, navathe2018hospital, hatef2019assessing} Developing high-performing information extraction models requires annotated data for supervised learning, and extraction performance is influenced by corpus size, heterogeneity, and annotation uniformity. 

\subsection*{Objective}

This paper summarizes the National NLP Clinical Challenges (n2c2) extraction task, \textit{Track 2: Extracting Social Determinants of Health} (\challengeName{}). The \challengeName{} utilized a novel corpus of annotated clinical text to evaluate the performance of a wide range of SDOH information extraction approaches from rules to deep learning language models (LM). The findings provide insight regarding extraction architecture development and learning strategies, generalizability to new domains, and remaining SDOH extraction challenges.


\ifsubfile
\bibliography{mybib}
\fi

\subsection*{Related Work}


SDOH are increasingly being recognized for their impact on health, and the body SDOH extraction research is also increasing.\cite{patra2021extracting} Clinical corpora have been annotated for a range of SDOH, including substance use, employment, living situation, environmental factors, physical activity, sexual factors, transportation, education, and language.\cite{LYBARGER2021103631, uzuner2008identifying, gehrmann2018comparing, feller2018towards, chapman2021rehoused, yu2021study, yu2022assessing, han2022classifying, wang2015automated, Yetisgen2017substance, reeves2021adaptation} Annotated SDOH data sets typically assign labels at the note-level or use a more granular relation-style structure. 

Many annotated SDOH data sets have sentence or note-level labels, approaching extraction as text classification.\cite{uzuner2008identifying, gehrmann2018comparing, feller2018towards, chapman2021rehoused, yu2021study, yu2022assessing, han2022classifying} The i2b2 NLP Smoking Challenge introduced a corpus of 502 notes with tobacco use status labels.\cite{uzuner2008identifying} Gehrmann et al. annotated 1,610 notes with phenotype labels, including substance abuse and obesity.\cite{gehrmann2018comparing} Feller et al. created a data set with 3,883 notes annotated for sexual health factors, substance use, and housing.\cite{feller2018towards} Chapman et al. annotated 621 notes for housing instability.\cite{chapman2021rehoused} Yu et al. created a corpus of 500 notes annotated with 15 SDOH concepts (marital status, education, occupation, etc.).\cite{yu2021study, yu2022assessing} Han et al. annotated 3,504 sentences for 13 SDOH including social environment, support networks, other factors.\cite{han2022classifying}

While most SDOH corpora use coarser note-level labels, some SDOH corpora utilize a more granular annotation scheme where spans and links between span are identified. Wang et al. annotated 691 notes using a relation-based scheme for substance use.\cite{wang2015automated} Yetisgen et al. annotated 364 notes using an event-based scheme for 13 SDOH (substance use, living situation, etc).\cite{Yetisgen2017substance} Reeves et al. annotated eight SDOH concepts (living situation, language, etc.) with assertion values (present vs. absent) in a corpus of 160 notes.\cite{reeves2021adaptation} 


SDOH information extraction techniques include rules, supervised learning, and unsupervised learning.\cite{patra2021extracting, bompelli2021social} Rules-based approaches include curated lexicons, regular expressions, and term expansion.\cite{patra2021extracting, hatef2019assessing, bompelli2021social, lowery2022using, reeves2021adaptation, chapman2021rehoused} Supervised extraction approaches include discrete input representations, like n-grams, term frequency-inverse document frequency (TF-IDF) , part-of-speech tags, and medical concepts. Discrete classification architectures include Support Vector Machines, random forest, logistic regression, maximum entropy, and conditional random fields.\cite{patra2021extracting, han2022classifying, wang2015automated, Yetisgen2017substance} Recent supervised extraction approaches utilize deep learning architectures, such as convolutional neural networks, recurrent neural networks, and transformers.\cite{patra2021extracting, gehrmann2018comparing, han2022classifying, yu2021study, yu2022assessing, LYBARGER2021103631}

\ifsubfile
\bibliography{mybib}
\fi


\section*{Materials and Methods}

\subsection*{Data}

\begin{figure}[ht]
    \centering
    \frame{\includegraphics[width=0.6\textwidth]{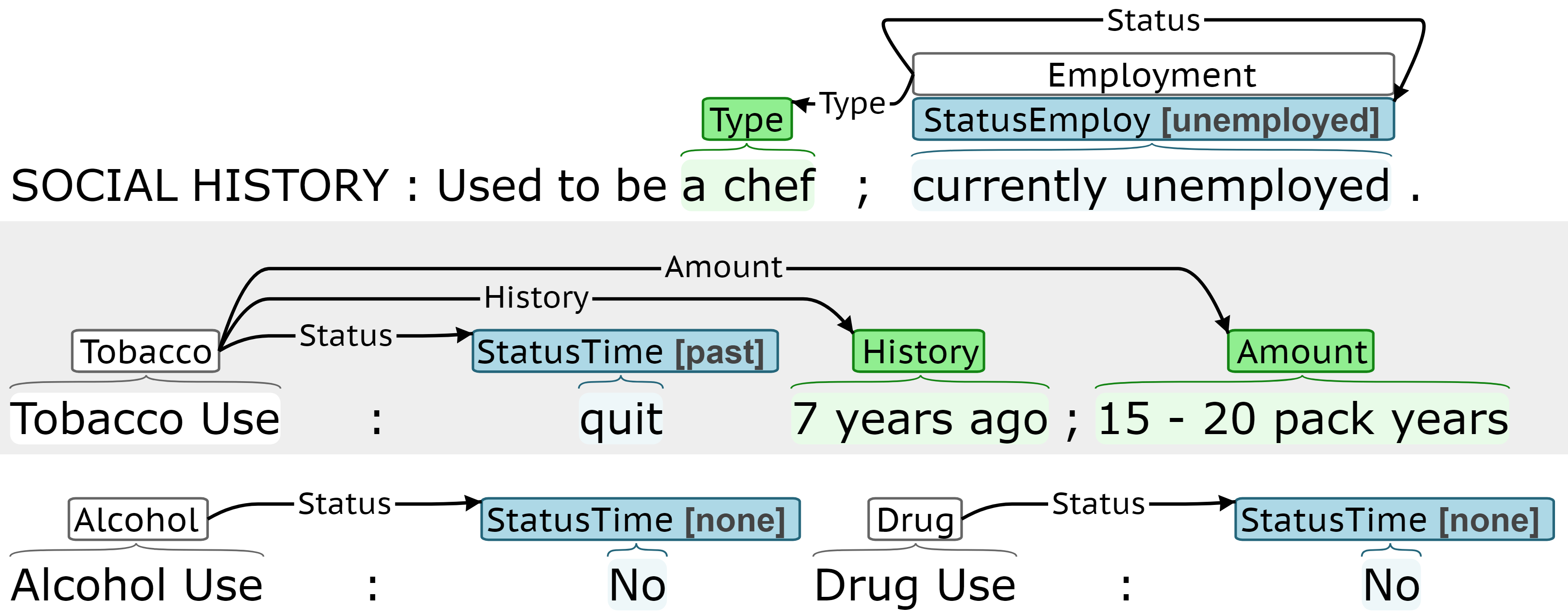}}
    \caption{BRAT annotation example.}
    \label{brat_example}
\end{figure}

The \challengeName{} used SHAC for model training and evaluation.\cite{LYBARGER2021103631} SHAC includes social history sections from clinical notes annotated for SDOH using an event-based scheme.\cite{LYBARGER2021103631} SHAC was annotated using the BRAT, which is available online\footnote{https://brat.nlplab.org/}.\citep{stenetorp2012brat} Figure 1 
presents an annotation example. Each event includes exactly one trigger (in white) and one or more arguments that characterize the event. There are two argument categories: \textit{span-only} (in green) and \textit{labeled} (in blue). The trigger anchors the event and indicates the event type (e.g. \textit{Employment}). \textit{Span-only arguments} include an annotated span and argument type (e.g. \textit{Duration}). \textit{Labeled arguments} include an annotated span, argument type (e.g. \textit{Status Time}), and argument subtype (e.g. \textit{past}). Argument roles connect triggers and arguments and can be interpreted as binary connectors, because there is one valid argument role type per argument type. For example, all \textit{Status Time} arguments connect to triggers through a \textit{Status} argument role. Table \ref{annotated_phenomena} summarizes the annotated phenomena. 

\begin{table}[ht]
    \small
    \centering

\begin{tabular}{p{0.7in} p{0.8in} p{2.2in} p{1.2in}}
\toprule
\textbf{Event type}                                         & \textbf{Argument type} & \textbf{Argument subtypes}                                                  & \textbf{Span examples}                                            \\ \midrule
\multirow{8}{0.8in}[\baselineskip]{Alcohol, Drug, \& Tobacco} & Status Time\textsuperscript{*}            & \{none, current, past\}                           & ``denies," ``smokes"                             \\ \cline{2-4} 
                                                            & Duration          & --                                                                  & ``for the past 8 years"                                            \\ \cline{2-4} 
                                                            & History           & --                                                                  & ``seven years ago"                                                 \\ \cline{2-4} 
                                                            & Type              & --                                                                  & ``beer," ``cocaine"                                         \\ \cline{2-4} 
                                                            & Amount            & --                                                                  & ``2 packs," ``3 drinks"                                           \\ \cline{2-4} 
                                                            & Frequency         & --                                                                  & ``daily," ``monthly"                                      \\ \hline 
\multirow{4}{0.8in}[\baselineskip]{Employment}                             & Status Employ\textsuperscript{*}            & \{\nohyphens{employed, unemployed, retired, \newline on disability, student, homemaker}\} & ``works," ``unemployed"               \\ \cline{2-4} 
                                                            & Duration          & --                                                                  & ``for five years"                                         \\ \cline{2-4} 
                                                            & History           & --                                                                  & ``15 years ago"                                                    \\ \cline{2-4} 
                                                            & Type              & --                                                                  & ``nurse," ``office work"                                 \\ \hline
\multirow{4}{0.8in}[\baselineskip]{Living Status}                          & Status Time\textsuperscript{*}            & \{current, past, future\}                            & ``lives," ``lived"                                         \\ \cline{2-4} 
                                                            & Type Living\textsuperscript{*}              & \{\nohyphens{alone, with family, with others, homeless}\}     & ``with husband," ``alone"                        \\ \cline{2-4} 
                                                            & Duration          & --                                                                  & ``for the past 6 months"                                         \\ \cline{2-4} 
                                                            & History           & --                                                                  & ``until a month ago"                                               \\ \bottomrule

\end{tabular}
    \caption{Annotation guideline summary. *indicates the argument is required.}
    \label{annotated_phenomena}
\end{table}

SHAC includes 4,405 social history sections from MIMIC-III and UW. MIMIC-III is a publicly available, deidentified health database for critical care patients at Beth Israel Deaconess Medical Center from 2001-2012.\citep{johnson2016mimic} SHAC samples were selected from 60K MIMIC-III discharge summaries. The UW data set includes 83K emergency department, 22K admit, 8K progress, and 5K discharge summary notes from the UW and Harborview Medical Centers generated between 2008-2019. We refer to these social history sections as \textit{notes}, even though each social history section is only part of the original note. Table \ref{shac_counts} summarizes the SHAC train, development, and test partitions. The development ($\mathcal{D}_{dev}$) and test ($\mathcal{D}_{test}$) partitions were randomly sampled. The train partition ($\mathcal{D}_{train}$) was 71\% actively selected, and the UW train partition ($\mathcal{D}_{train}^{uw}$) was 79\% actively selected using a novel active learning framework. The active learning framework selected batches for annotation based on diversity and informativeness. To assess diversity, samples were mapped into a vector space using TF-IDF weighted averages of pre-trained word embeddings. To assess informativeness, simplified note-level text classification tasks for substance use, employment, and living status were derived from the event annotations and used as a surrogate for event extraction. Sample informativeness was assessed as the entropy of note-level predictions for: \textit{Status Time} for \textit{Alcohol}, \textit{Drug}, and \textit{Tobacco}; \textit{Type Employ} for \textit{Employment}; and \textit{Status Living} for \textit{Living Status}. These labels capture normalized representations of social protective and risk factors. Active learning increased sample diversity and the prevalence of risk factors, like housing instability and polysubstance use. It improved extraction performance, with the largest performance gains associated with important risk factors, like drug use, homelessness, and unemployment.\cite{LYBARGER2021103631}

\begin{table}[ht]
    \small
    \centering

\begin{tabular}{llll  l}
\toprule
\textbf{Source} & \textbf{Train}                                    & \textbf{Dev}                        & \textbf{Test}                        & \textbf{Total} \\ \midrule
MIMIC-III       & 1,316   ($\mathcal{D}_{train}^{mimic}$) & 188 ($\mathcal{D}_{dev}^{mimic}$) & 373 ($\mathcal{D}_{test}^{mimic}$) & 1,877 ($\mathcal{D}^{mimic}$)  \\
UW              & 1,751   ($\mathcal{D}_{train}^{uw}   $) & 259 ($\mathcal{D}_{dev}^{uw}   $) & 518 ($\mathcal{D}_{test}^{uw}   $) & 2,528 ($\mathcal{D}^{uw}   $)  \\  \midrule
Total           & 3,067   ($\mathcal{D}_{train}        $) & 447 ($\mathcal{D}_{dev}        $) & 891 ($\mathcal{D}_{test}        $) & 4,405 ($\mathcal{D}$)          \\ \bottomrule
\end{tabular}
    \caption{SHAC note counts by source.}
    \label{shac_counts}
\end{table}

To prepare it for release, SHAC ($\mathcal{D}$) was reviewed, and annotations were adjusted to improve uniformity. The UW partition ($D^{uw}$) was deidentified using an in-house deidentification model \cite{lee2021transferability} and through manual review of each note by three annotators. Most protected health information (PHI) was replaced with special tokens related to age, contact information, dates, identifiers, locations, names, and professions (e.g. ``[DATE]''). However, the meaning of some annotations relies on spans identified as PHI. To preserve annotation meaning, some PHI spans were replaced with randomly selected surrogates from curated lists. For example, named homeless shelters (e.g. ``Union Gospel Mission'') are important to the \textit{Type Living} label \textit{homeless}, and named shelters were replaced with random selections from a list of regional shelters. Surrogates were also manually created to reduce specificity. For example, an \textit{Employment} \textit{Type} span, like ``UPS driver,'' may be replaced with ``delivery driver.'' 

\subsection*{Subtasks}
The challenge includes three subtasks:

\begin{itemize}
    \item \textit{Subtask A (Extraction)} focuses on in-domain extraction, where the training and evaluation data are from the same domain. The training data include the MIMIC-III train and development partitions ($\mathcal{D}_{train}^{mimic}, \mathcal{D}_{dev}^{mimic}$), and the evaluation data include the MIMIC-III test partition ($\mathcal{D}_{test}^{mimic}$).
    \item \textit{Subtask B (Generalizability)} explores generalizability to an unseen domain, where the training and evaluation data are from different domains. The training data are the same as Subtask A, and the evaluation data are the UW train and development partitions ($\mathcal{D}_{train}^{uw}, \mathcal{D}_{dev}^{uw}$).
    \item \textit{Subtask C (Learning Transfer)} investigates learning transfer, where the training data include in-domain and out-domain data. The training data are the MIMIC-III and UW train and development partitions ($\mathcal{D}_{train}^{mimic}$, $\mathcal{D}_{dev}^{mimic}$, $\mathcal{D}_{train}^{uw}$, $\mathcal{D}_{dev}^{uw}$), and the evaluation data are the UW test partition ($\mathcal{D}_{test}^{uw}$).	 
\end{itemize}


\subsection*{Shared Task Structure}
The \challengeName{} was conducted in 2022. To access SHAC, participants were credentialed through PhysioNet for MIMIC-III \cite{johnson2016mimic} and submitted data use agreements for the UW partition. For Subtasks A and B, training data were provided on February 14, unlabeled evaluation data were provided on June 6, and system predictions were submitted by participants on June 7. For Subtask C, training data were released on June 8, unlabeled evaluation data were released on June 9, and system predictions were submitted by participants on June 10. Teams were allowed to submit three sets of predictions for each subtask, and the highest performing submission for each subtask was used to determine team rankings.

\subsection*{Scoring}

Extraction requires identification of trigger and argument spans, resolution of trigger-argument connections, and prediction of argument subtypes. The evaluation criteria interprets the extraction task as a slot filling task, as this is most relevant to secondary use applications.  Figure 2 
presents the same sentence with two sets of annotations, \textit{A} and \textit{B}, along with the populated slots. Both annotations identify two \textit{Drug} events: \textit{Event 1} and \textit{Event 2}. Event 1 describes past intravenous drug use (IVDU), and Event 2 describes current cocaine use. Event 1 is annotated identically by \textit{A} and \textit{B}. However, there are differences in the annotated spans of Event 2, specifically for the trigger (``cocaine" vs. ``cocaine use") and \textit{Status Time} (``use" vs. ``Recent"). From a slot perspective, the annotations for Event 2 could be considered equivalent. 
The evaluation uses relaxed criteria for triggers and labeled arguments that reflect the clinical meaning of the extraction. The criteria and justification are presented below.

\begin{figure}[H]
    \centering
    \frame{\includegraphics[width=0.9\textwidth]{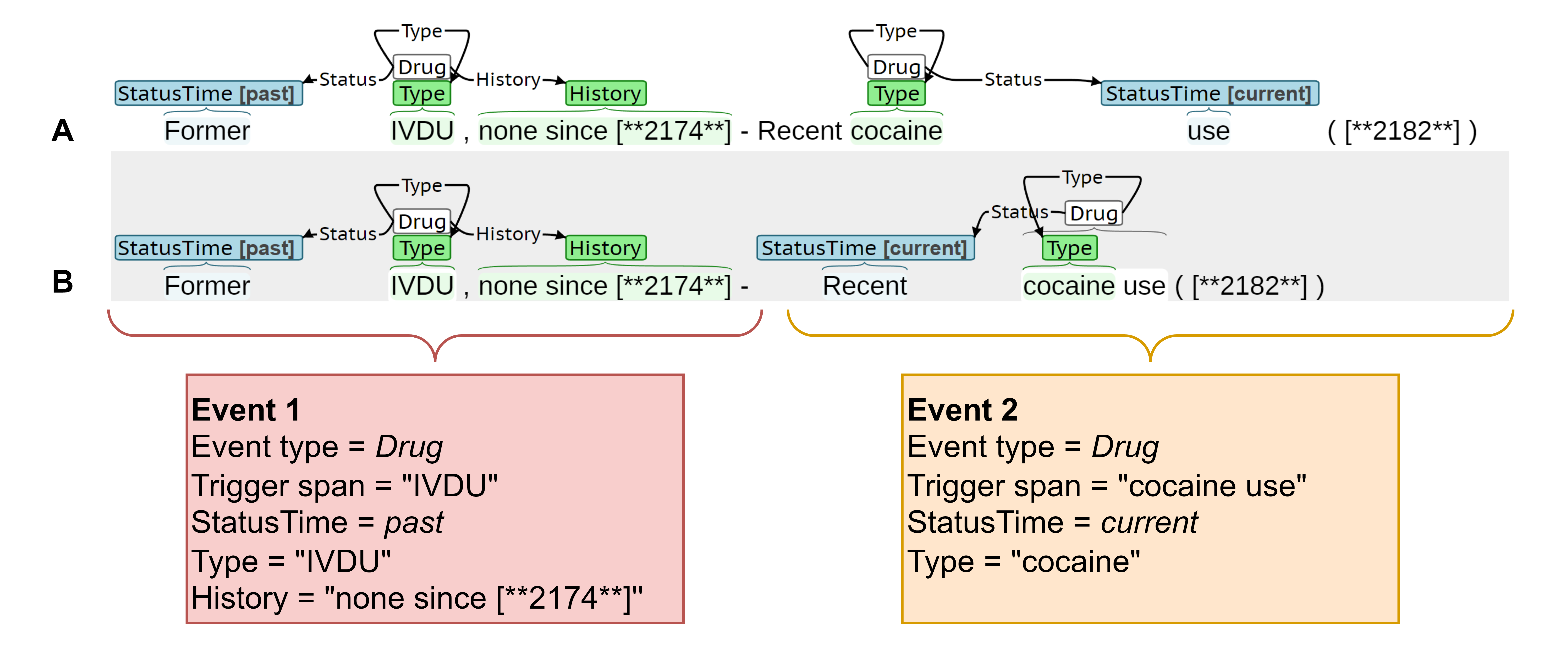}}
    \caption{Annotation examples describing event extraction as a slot filling task.}
    \label{annotation_comparison}
\end{figure}

\textit{Trigger:} A trigger is defined by an event type (e.g. \textit{Drug}) and multi-word span (e.g. phrase ``cocaine use'').  In SHAC, the primary function of the trigger is to anchor the event and aggregate related arguments. The text of the trigger span typically does not meaningfully contribute to the event meaning. Consequently, trigger equivalence is defined using \textit{any overlap} criteria where two triggers are equivalent if: 1) the event types match and 2) the spans overlap by at least one character. For \textit{Event 2} in Figure 2
, the trigger in \textit{A} has the event type \textit{Drug} and span ``cocaine,'' and the trigger in \textit{B} has the event type \textit{Drug} and span ``cocaine use.'' These triggers are equivalent under this \textit{any overlap} criteria.


\textit{Arguments:} Events are aligned based on trigger equivalence, and the arguments of aligned events are compared using different criteria for \textit{span-only} and \textit{labeled arguments}. 

\textit{Span-only arguments:}  A span-only argument is defined by an argument type (e.g. \textit{Type}), span (e.g. phrase ``cocaine''), and connection to a trigger. Span-only argument equivalence uses \textit{exact match} criteria, where two span-only arguments are equivalent if: 1) the connected triggers are equivalent, 2) the argument types match, and 3) the spans match exactly.



\textit{Labeled arguments:} A labeled argument is defined by an argument type (e.g. \textit{Status Time}), argument subtype (e.g. \textit{current}), argument span (e.g. phrase ``Recent''), and connection to a trigger. The argument subtypes normalize the span text and capture a majority of the span semantics. There was often ambiguity in the labeled argument span annotation. To focus the evaluation on the most salient information (subtype labels) and address the ambiguity in argument span annotation, labeled argument equivalence is defined using a \textit{span agnostic} approach, where two labeled arguments are equivalent if: 1) the connected triggers are equivalent, 2) the argument types match, and 3) the argument subtypes match. 




Performance is evaluated using precision (P), recall (R), and F1, micro averaged over the event types, argument types, and argument subtypes. Submissions were compared using an overall F1 score calculated by summing the true positives, false-negatives, and false-positives across all annotated phenomena. The scoring routine is available\footnote{https://github.com/Lybarger/brat\_scoring}. Significance testing was performed on the overall F1 scores using a paired bootstrap test with 10,000 repetitions, where samples were selected at the note-level.

\subsection*{Systems}

\begin{table}[ht]
    \small
    \centering

\begin{tabular}{p{1.8in} p{0.45in} p{0.7in} p{0.7in} p{0.45in}  p{0.5in}}
\toprule
\textbf{Team (in alphabetical order)}                            & \textbf{Team Abbrv}. &  \textbf{Language Rep.} & \textbf{Architecture} & \textbf{External Data}       &  \textbf{Subtasks} \\ \midrule  
Children's Hospital of Philadelphia$^*$  & CHOP                 &  LM$^c$                 & ST+TC                 & --                           &  A, B, C  \\ 
IBM$^*$                                  & IBM                  &  LM$^c$                 & unknown               & --                           &  C        \\ 
Kaiser Permanente Southern CA$^*$        & KP                   &  LM$^{c,b}$             & ST+TC                 & --                           &  A, B     \\ 
Medical Univ. of South Carolina$^*$      & MUSC                 &  n-grams                & KB$^u$, rules         & --                           &  A, B, C  \\ 
Microsoft$^*$                            & MS                   &  LM$^t$                 & seq2seq               & $\mathcal{U}$, $\mathcal{L}$ &  A, B, C  \\ 
Philips Research North America$^*$       & PR                   &  LM$^t$                 & seq2seq               & --                           &  A, B, C  \\ 
University Medical Center Utrecht$^*$    & UMCU                 &  LM$^i$, WE             & ST+TC                 & $\mathcal{U}$                &  A, B, C  \\ 
University of Florida$^*$                & UFL                  &  LM$^i$                 & ST+TC                 & $\mathcal{U}$                &  A, B, C  \\ 
University of Massachusetts              & UMass                &  unknown                & unknown               & --                           &  A, C     \\ 
University of Michigan$^*$               & UM                   &  n-grams                & ST+TC, rules          & --                           &  A, B     \\ 
University of New South Wales            & UNSW                 &  n-grams, WE            & ST+TC, rules          & --                           &  A, B     \\ 
University of Pittsburgh                 & Pitt                 &  n-grams                & rules                 & --                           &  A        \\   
University of Texas at San Antonio$^*$   & UTSA                 &  LM$^r$, WE             & ST+TC                 & --                           &  A, B, C  \\ 
University of Utah$^*$                   & UU                   &  LM$^c$                 & ST+TC                 & --                           &  A, C     \\  
Verily Life Sciences                     & Verily               &  LM$^c$                 & ST+TC                 & --                           &  A        \\  \bottomrule
\end{tabular}

    \caption{Summary of top performing system for each team. $^*$ indicates the team submitted an abstract. $^c$ indicates ClinicalBERT.\cite{alsentzer-etal-2019-publicly} $^b$ indicates BioBERT.\cite{lee2019biobert} $^t$ indicates T5.\cite{raffel2020exploring} $^i$ indicates an in-house BERT variant pretrained on institutional data. $^u$ indicates the Unified Medical Language System (UMLS) \cite{bodenreider2004unified}. $^r$ indicates RoBERTa \cite{zhuang2021robustly}}
    \label{systems}
\end{table}

This section summarizes the methodologies of the participating teams. Table \ref{systems} summarizes the best performing system for each team, including language representation, architecture, and external data. Language representation includes: i) \textit{n-grams} - discrete word representations, ii) \textit{pretrained word embeddings (WE)} - vector representation of words, like word2vec,\cite{Mikolov_2013_word2vec} and iii) \textit{pretrained language models (LM)} - deep learning LM, like BERT \cite{devlin2019bert} and T5 \cite{raffel2020exploring}. Architecture includes: i) \textit{rules} - hand-crafted rules, ii) \textit{knowledge based (KB)} - dictionaries and ontologies, iii) \textit{sequence tagging + text classification (ST+TC)} - combination of sequence tagging and text classification layers, and iv) \textit{sequence-to-sequence (seq2seq)} - text generation models, like T5, transform unstructured input text into a structured representation. External data includes: i) \textit{unlabeled ($\mathcal{U}$)} - unsupervised learning with unlabeled data and ii) \textit{labeled ($\mathcal{L}$)} - supervised learning with labeled data other than SHAC. Many teams used pretrained WE and LM; however, $\mathcal{U}$ is only assigned for additional pretraining, beyond publicly available models. For example, ClinicalBERT \cite{alsentzer-etal-2019-publicly} would not be assigned $\mathcal{U}$; however, additional pretraining of ClinicalBERT would be assigned $\mathcal{U}$.

Below is a summary of the teams that achieved first and second place in each subtask. The top teams were determined based on the performance in Table \ref{results} and are presented below alphabetically.
\begin{itemize}
    \item \textit{Children's Hospital of Philadelphia (CHOP)} designed a BERT-based pipeline consisting of trigger identification and argument resolution. Triggers were extracted as a sequence tagging task using BERT with a linear prediction layer at the output hidden states.  For argument extraction, a unique input representation was created for each identified trigger, where the segment IDs were 1 for the target trigger and 0 elsewhere. Arguments were identified using multiple linear layers applied to the BERT hidden states, to allow multi-label token predictions. Encoding the target trigger using the segment IDs allowed BERT to focus the argument prediction on a single event. Argument subtype labels were resolved by creating separate tags for each subtype (e.g., ``LivingStatus=alone''). 
    \item \textit{Kaiser Permanente Southern CA (KP)} assumed there is at most one event per event type per sentence, similar to the original SHAC paper. Trigger and labeled argument prediction was performed as a binary text classification task (e.g. no alcohol vs. has alcohol; no current alcohol use vs. current alcohol use) using BERT with a linear layer applied to the pooled state. Trigger and span-only arguments spans were extracted using BERT with a linear layer applied to the hidden states, with separate models for each. Within a sentence, the extracted trigger and arguments for a given event type were assumed to be part of the same event. The trigger-argument linking was implicit in the assumption that there is at most one event per SDOH per sentence.
    \item \textit{Microsoft (MS)} developed a seq2seq approach with T5-large as the pretrained encoder-decoder. T5 was further pretrained on MIMIC-III notes. In fine-tuning, the input was note text, and the output was a structured text representation of SDOH. Additional negative samples were incorporated using MIMIC-III text that does not include SDOH, and additional positive samples were created from the annotated data by excerpting shorter samples ($\approx9$ new samples per note). In addition to the challenge data, MS used an in-house data set with SDOH annotations; however, an MS ablation study indicated this in-house data had a marginal impact on performance. A constraint solver post-processed the structured text output to generate the text offsets for the final span predictions.
    \item \textit{University of Florida (UFL)} designed a multi-step approach: 1) span extraction, 2) relation prediction and 3) argument subtype classification. Trigger and argument types were divided into five groups, where overlapping spans are minimized within each group. The trigger and argument spans for each group were then extracted using a separate BERT model with a linear layer at the hidden states. Relation prediction was a binary text classification task using BERT, where the input included special tokens demarking target trigger and argument spans. Argument subtype classification was similar to relation prediction, except the targets were the subtype labels. UFL used an internal BERT variant, GatorTron, which was trained on in-house and public data. 
\end{itemize}

\ifsubfile
\bibliography{mybib}
\fi

\section*{Results}

\begin{table}[!ht]
    
    \newcommand{\team}{\multirow{2}{*}{\textbf{Team}}}
    \newcommand{\precision}{\multirow{2}{*}{\textbf{P}}}
    \newcommand{\recall}{\multirow{2}{*}{\textbf{R}}}
    \newcommand{\fscore}{\multirow{2}{*}{\textbf{F1}}}
    \newcolumntype{C}[1]{>{\centering\arraybackslash}p{#1}}

    \newcommand{\widthT}{0.50in}    
    \newcommand{\widthN}{0.35in}
    \newcommand{\widthS}{0.25in}
    
    \centering
    \small
    
    \begin{subtable}[h]{1.0\textwidth}
        \centering

\begin{tabular}{p{\widthT}  C{\widthN} C{\widthN} C{\widthN}  C{\widthS}  C{\widthS}  C{\widthS}  C{\widthS}  C{\widthS}  C{\widthS}  C{\widthS}  C{\widthS} }

\toprule
\team   & \precision & \recall & \fscore & \multicolumn{8}{c} {\textbf{Significance}}\\  \cline{5-12} 
        &            &         &         & \textbf{MS} & \textbf{UFL} & \textbf{KP} & \textbf{CHOP} & \textbf{PR} & \textbf{UTSA} & \textbf{UMCU} & \textbf{Verily}\\ \midrule
MS      &  0.909     & 0.893   & 0.901   & --          & \xmark       & \cmark      & \cmark        & \cmark      & \cmark        & \cmark        & \cmark         \\ \hline
UFL     &  0.878     & 0.908   & 0.893   & --          & --           & \xmark      & \cmark        & \cmark      & \cmark        & \cmark        & \cmark         \\ \hline
KP      &  0.884     & 0.884   & 0.884   & --          & --           & --          & \xmark        & \cmark      & \cmark        & \cmark        & \cmark         \\ \hline
CHOP    &  0.870     & 0.889   & 0.879   & --          & --           & --          & --            & \cmark      & \cmark        & \cmark        & \cmark         \\ \hline
PR      &  0.862     & 0.842   & 0.852   & --          & --           & --          & --            & --          & \xmark        & \cmark        & \cmark         \\ \hline
UTSA    &  0.847     & 0.827   & 0.837   & --          & --           & --          & --            & --          & --            & \cmark        & \cmark         \\ \hline
UMCU    &  0.864     & 0.653   & 0.744   & --          & --           & --          & --            & --          & --            & --            & \cmark         \\ \hline
Verily  &  0.797     & 0.529   & 0.636   & --          & --           & --          & --            & --          & --            & --            & --             \\ \bottomrule
\end{tabular}


        \caption{Subtask A}
        \label{results_a}
    \end{subtable}
    
    \vspace{0.125in}

    \begin{subtable}[h]{1.0\textwidth}
        \centering

\begin{tabular}{p{\widthT}  C{\widthN} C{\widthN} C{\widthN}  C{\widthS}  C{\widthS}  C{\widthS}  C{\widthS}  C{\widthS}  C{\widthS}  C{\widthS}  C{\widthS} }

\toprule
\team           & \precision & \recall & \fscore & \multicolumn{8}{c} {\textbf{Significance}}\\  \cline{5-12} 
                &            &         &         & \textbf{MS} & \textbf{KP}  & \textbf{CHOP} & \textbf{UFL}& \textbf{UTSA} & \textbf{PR} & --            & -- \\ \midrule
MS              &  0.811     & 0.740   &  0.774  & --          & \xmark       & \cmark      & \cmark        & \cmark      & \cmark        & --            & --             \\ \hline
KP              &  0.790     & 0.748   &  0.768  & --          & --           & \xmark      & \cmark        & \cmark      & \cmark        & --            & --             \\ \hline
CHOP            &  0.761     & 0.770   &  0.766  & --          & --           & --          & \cmark        & \cmark      & \cmark        & --            & --             \\ \hline
UFL             &  0.761     & 0.713   &  0.736  & --          & --           & --          & --            & \cmark      & \cmark        & --            & --             \\ \hline
UTSA            &  0.737     & 0.665   &  0.699  & --          & --           & --          & --            & --          & \xmark        & --            & --             \\ \hline
PR              &  0.739     & 0.655   &  0.695  & --          & --           & --          & --            & --          & --            & --            & --             \\ \bottomrule
\end{tabular}

        \caption{Subtask B}
        \label{results_b}
    \end{subtable}

    \vspace{0.125in}

    \begin{subtable}[h]{1.0\textwidth}
        \centering

\begin{tabular}{p{\widthT}  C{\widthN} C{\widthN} C{\widthN}  C{\widthS}  C{\widthS}  C{\widthS}  C{\widthS}  C{\widthS}  C{\widthS}  C{\widthS}  C{\widthS} }

\toprule
\team         & \precision & \recall & \fscore     & \multicolumn{8}{c} {\textbf{Significance}}\\  \cline{5-12} 
              &            &         &             & \textbf{MS} & \textbf{CHOP} & \textbf{UTSA} & \textbf{RP} & \textbf{UFL} & \textbf{UMCU} & \textbf{IBM} & -- \\ \midrule
MS            & 0.891      & 0.887   & 0.889       & --          & \xmark       & \cmark      & \cmark        & \cmark      & \cmark        & \cmark        & --             \\ \hline
CHOP          & 0.874      & 0.888   & 0.881       & --          & --           & \cmark      & \cmark        & \cmark      & \cmark        & \cmark        & --             \\ \hline
UTSA          & 0.880      & 0.841   & 0.860       & --          & --           & --          & \cmark        & \cmark      & \cmark        & \cmark        & --             \\ \hline
PR            & 0.885      & 0.780   & 0.829       & --          & --           & --          & --            & \cmark      & \cmark        & \cmark        & --             \\ \hline
UFL           & 0.923      & 0.683   & 0.785       & --          & --           & --          & --            & --          & \cmark        & \cmark        & --             \\ \hline
UMCU          & 0.886      & 0.553   & 0.681       & --          & --           & --          & --            & --          & --            & \cmark        & --             \\ \hline
IBM           & 0.538      & 0.788   & 0.640       & --          & --           & --          & --            & --          & --            & --            & --             \\ \bottomrule
\end{tabular}

        \caption{Subtask C}
        \label{results_c}
    \end{subtable}    
    
    \caption{Performance of top performing teams. Significance evaluated for F1 score at $p < 0.05$. \cmark{ }indicates the systems \textit{were} statistically different, and \xmark{ }indicates the system \textit{were not} statistically different.}
    
    \label{results}
\end{table}

Table \ref{results} presents the overall performance for Subtasks A, B, and C, including significance testing. Table \ref{results} includes the results from teams with F1 greater than the mean across all teams. Since the original SHAC publication, the UW SHAC partition was deidentified, and the entirety of SHAC was reviewed to improve annotation consistency. Additionally, the scoring routine used in the challenge differs from the original SHAC publication. Consequently, the results in Table \ref{results} should be considered the current state-of-the-art for SHAC. MS achieved the highest overall F1 across all three subtasks; however, the MS performance was not statistically different from the second team in each subtask: UFL for Subtask A, KP for Subtask B, and CHOP for Subtask C. Performance was substantively lower for Subtask B than Subtasks A and C. No in-domain training data was used in Subtask B. Additionally, the Subtask A and Subtask C evaluation data were randomly selected; however, the evaluation data for Subtask B included actively selected samples that intentionally biased the data towards risk factors. 

\subsection*{Error Analysis}

\label{errorAnalysis}

\begin{figure}[!ht]
    \centering
    \frame{\includegraphics[width=6.0in]{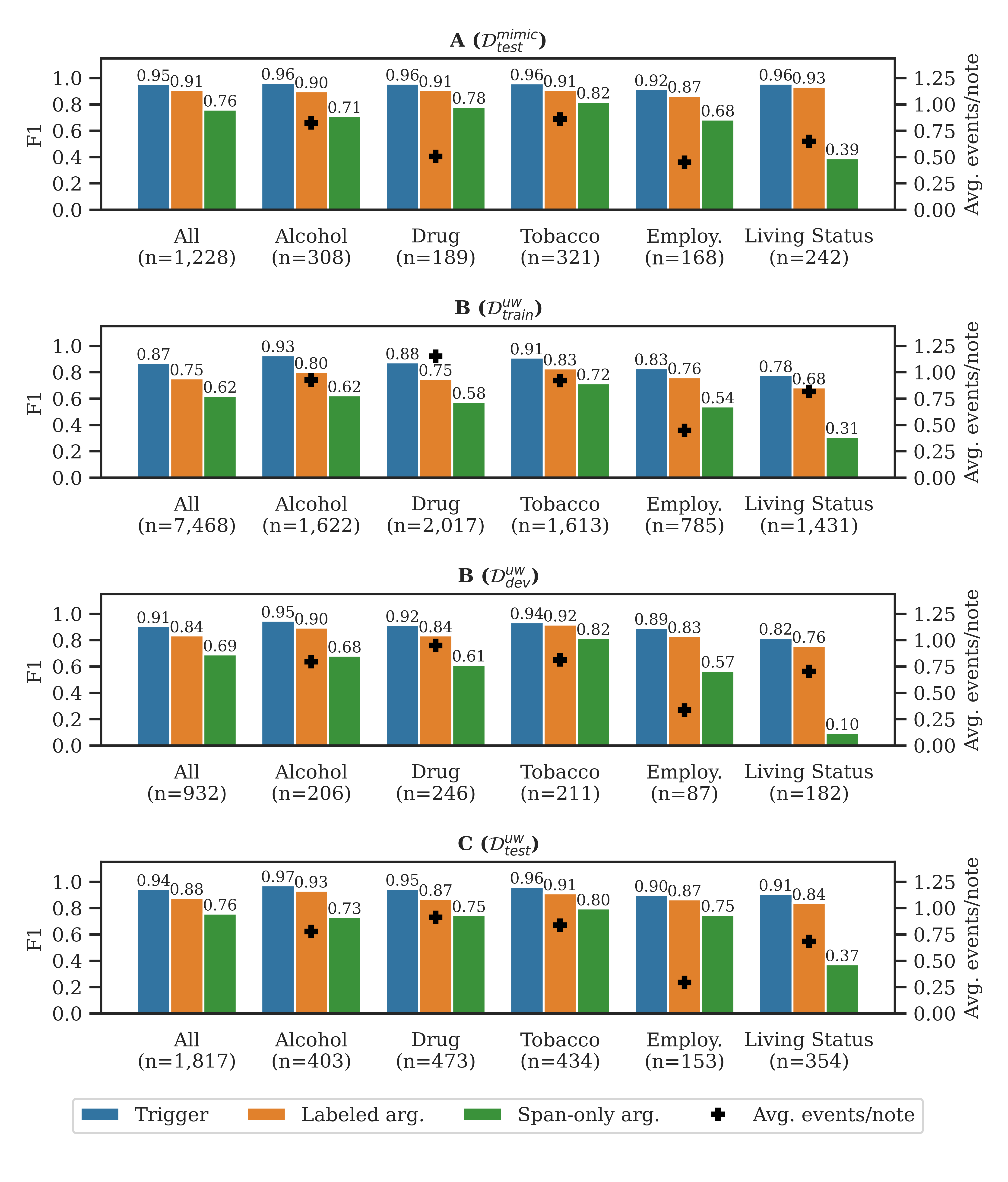}}
    \caption{Performance for top performing teams in Subtasks A, B, and C. The left-hand y-axis is the micro-averaged F1 for triggers, labeled arguments, and span-only arguments (vertical bars). The right-hand y-axis is the average number of gold events per note (\ding{58}).}
    \label{subtask_comparison}
\end{figure}

\begin{figure}[!ht]
    \centering
    \frame{\includegraphics[width=6in]{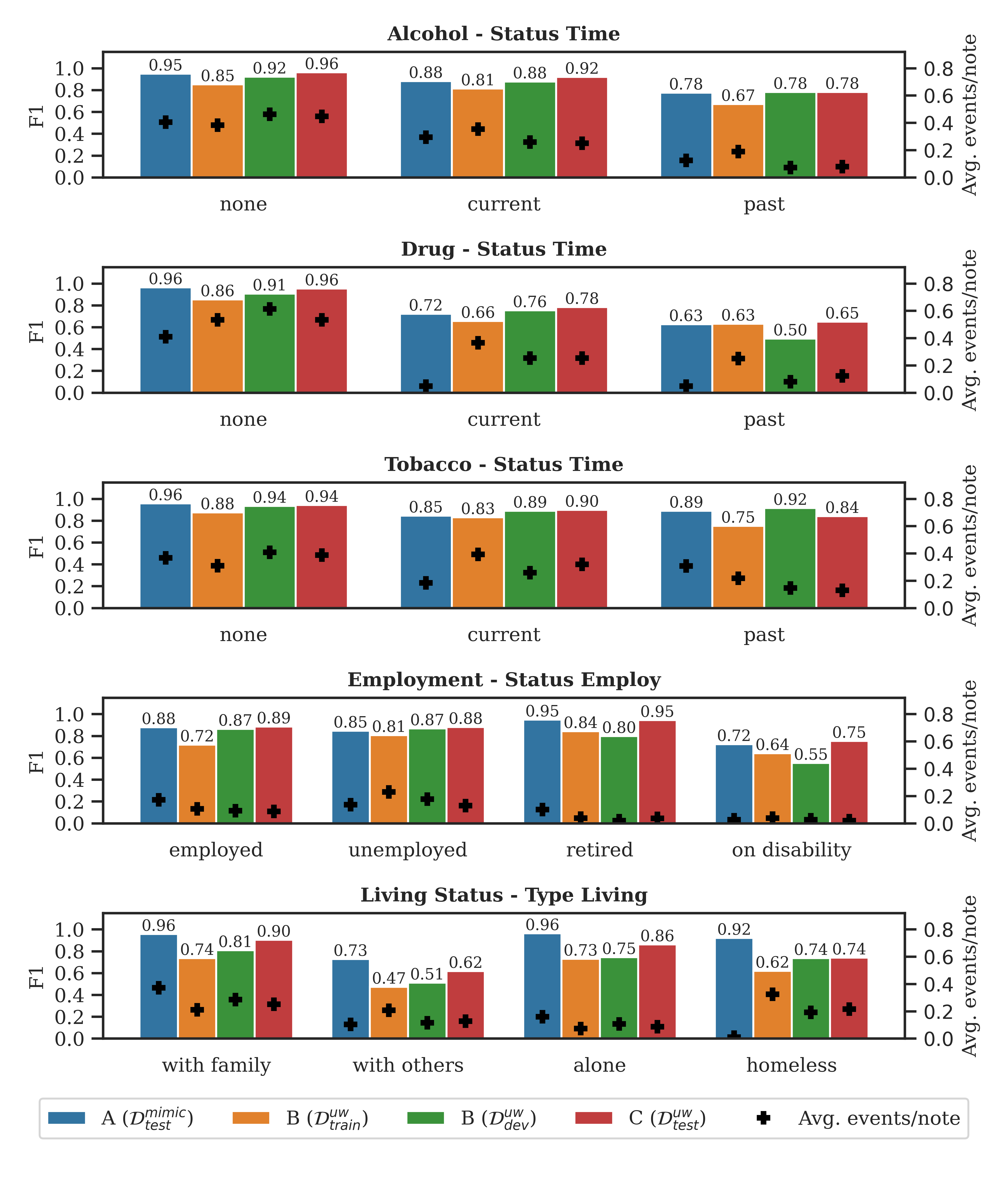}}
    \caption{Performance breakdown by argument subtype label for Subtasks A, B, and C. The left-hand y-axis is the micro-averaged F1 for the argument subtype labels (vertical bars). The right-hand y-axis is the average number of gold events per note (\ding{58}).}
    \label{risk_factor}
\end{figure}

\begin{figure}[!ht]
    \centering
    \frame{\includegraphics[width=6in]{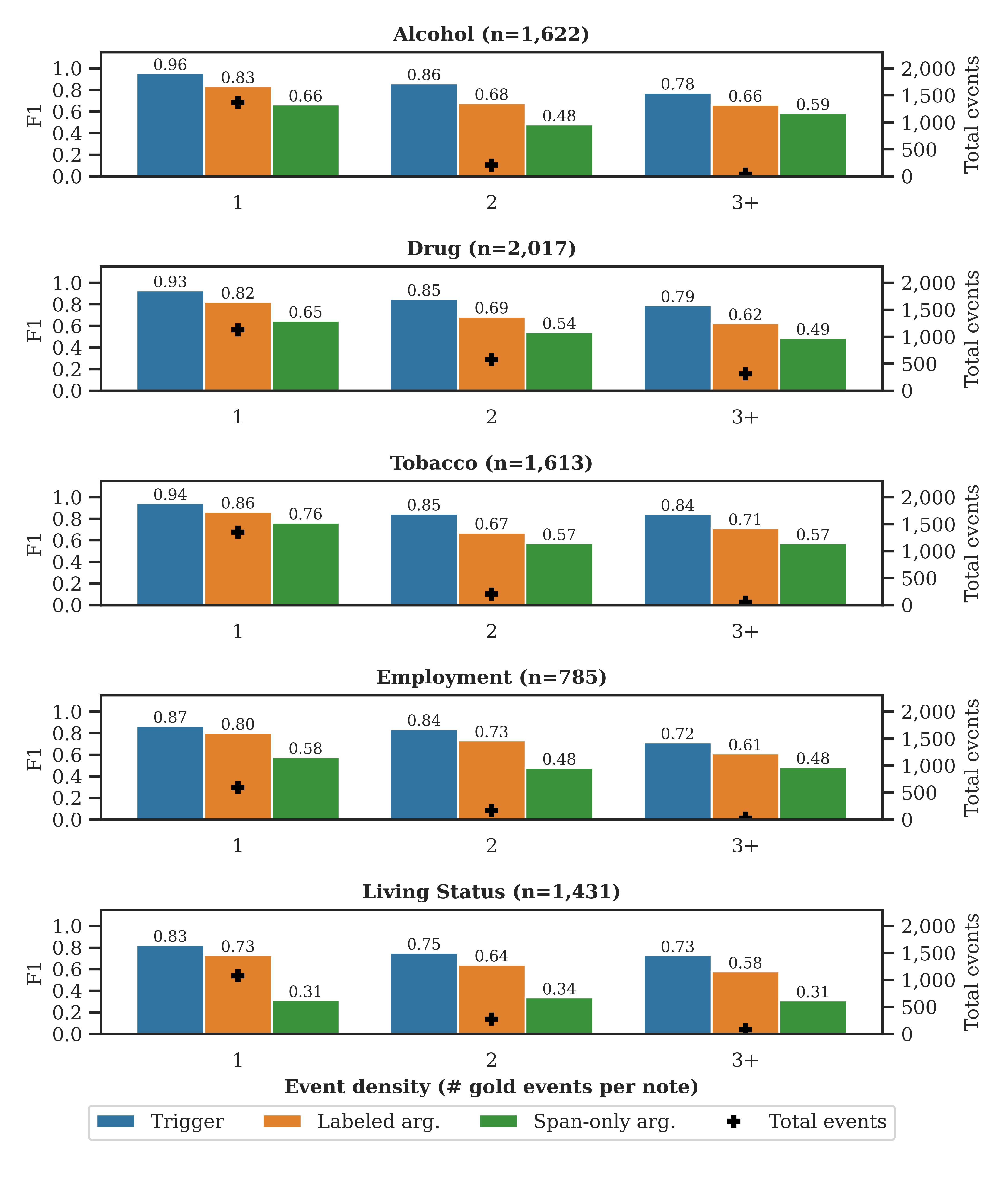}}
    \caption{Performance breakdown by event density for Subtask B ($\mathcal{D}_{train}^{uw}$). The left-hand y-axis is the micro-averaged F1 for triggers, labeled arguments, and span-only arguments (vertical bars). The right-hand y-axis is the total number of gold events (\ding{58}).}
   \label{density}
\end{figure}

We explored the performance by event type, argument subtype, and event frequency per note. The error analysis presents the micro-averaged performance of the top three teams in each subtask (Subtask A - MS, UFL, and KP; Subtask B - MS, KP, and CHOP; and Subtask C - MS, CHOP, and UTSA). Subtask B performance is presented separately for $\mathcal{D}_{train}^{uw}$ and $\mathcal{D}_{dev}^{uw}$, because of the differing sampling strategies (active vs. random).  

Figure 3 
presents the extraction performance by subtask and includes the total number of gold events ($n$) and average number of gold events per note ({\small \ding{58}}). Subtask A focused on in-domain extraction, and Subtask B focused on generalizability. The performance drops from Subtask A to Subtask B ($\mathcal{D}_{dev}^{uw}$) at {-0.04 $\Delta$F1} for triggers, {-0.07 $\Delta$F1} for labeled arguments, and {-0.07 $\Delta$F1} span-only arguments (\textit{All} category in figure). The largest performance drop is associated with \textit{Living Status} at {-0.14 $\Delta$F1} for triggers and {-0.17 $\Delta$F1} for labeled arguments. Subtask C explored learning transfer, where both in-domain and out-domain data are available. Comparing Subtask B ($\mathcal{D}_{dev}^{uw}$) and Subtask C, incorporating in-domain data increased performance, such that performance for Subtasks A and C are similar. The performance difference between $\mathcal{D}_{train}^{uw}$ and $\mathcal{D}_{dev}^{uw}$ in Subtask B demonstrates the actively selected samples are more challenging extraction targets.

SHAC annotations captures normalized SDOH factors through the arguments: \textit{Status Time} for substance use, \textit{Status Employ} for \textit{Employment}, and \textit{Type Living} for \textit{Living Status}. Figure 4 
presents the performance for these factors and the average number of gold events per note ({\small \ding{58}}). The \textit{student} and \textit{homemaker} labels for \textit{Status Employ} are omitted, due to low frequency. For substance use, the \textit{none} label has the highest performance where descriptions tend to be relatively concise and have low linguistic variability (e.g. ``Tobacco use: none’’ or ``Denies EtOH’’). Performance is lower for \textit{current} and \textit{past} subtype labels, which are associated with more linguistic diversity and have higher label confusability. Performance is lower for \textit{Drug} than \textit{Alcohol} and \textit{Tobacco} because of the more heterogeneous descriptions of drug use and types (e.g. ``cocaine remote,’’ ``smokes MJ,’’ and ``injects heroin’’). Regarding \textit{Employment}, the \textit{retired} label has the highest performance, due to the consistent use of the keyword ``retired.'' The \textit{employed} and \textit{unemployed} labels have similar performance that is incrementally less than \textit{retired}. The \textit{on disability} label has the lowest performance, which is partially attributable to annotation inconsistency and classification confusability associated with disambiguating the presence of disability and receiving disability benefits (e.g. `She does not work because of her disability'' vs. ``On SSI disability'').

Figure 5 
presents the performance for Subtask B ($\mathcal{D}_{train}^{uw}$) by the number of gold events per note for a given event type (referred to as \textit{event density}): one (1), two (2), and three or more (3+) events per note. Figure 5 
also includes the total number of gold events ({\small \ding{58}}). Across event types, performance is lower for multi-event notes (density $>$ 1) than single-event notes (density $=$ 1). Multi-event notes tend to have more detailed and nuanced options of the SDOH. All aspects of the extraction task, including span identification, span linking, and argument subtype resolution, are more challenging in multi-event notes. The appendix includes similar figures for Subtask A, Subtask B ($\mathcal{D}_{dev}^{uw}$), and Subtask C. Subtask B ($\mathcal{D}_{train}^{uw}$) is presented here because it has the highest event density.

\begin{figure}[ht]
    \centering

        \begin{subfigure}[b]{6in}
            \framebox{
                \begin{minipage}[t]{5.75in}
                    \includegraphics[scale=0.30]{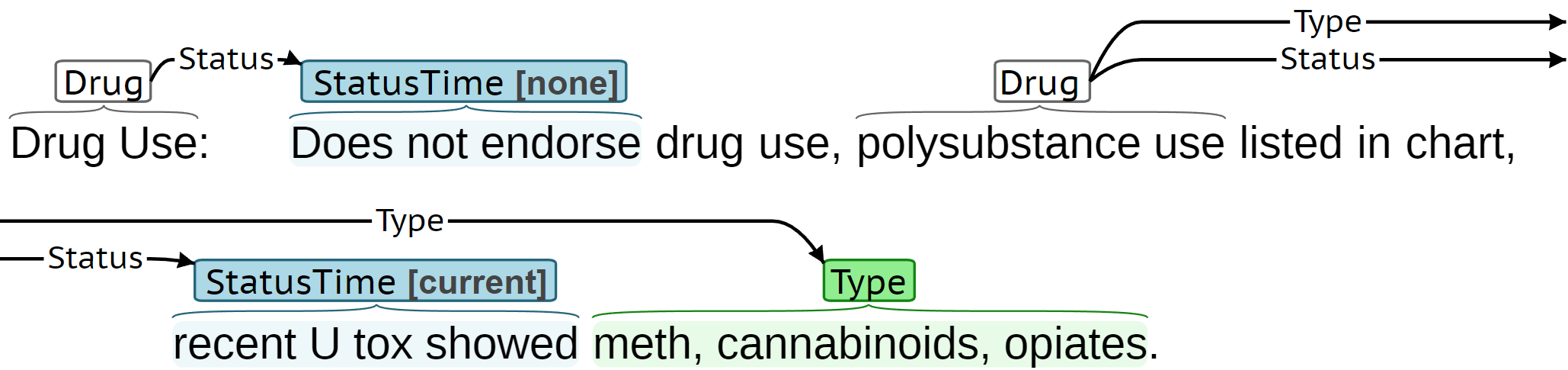}
                \end{minipage}
            }
            \caption{Multiple gold \textit{Drug} events in a sentence with conflicting \textit{Status Time} labels.}
            \label{error_analysis_examples_a}
        \end{subfigure}
        
       \begin{subfigure}[b]{6in}
           \framebox{
               \begin{minipage}[t]{5.75in}
                   \includegraphics[scale=0.30]{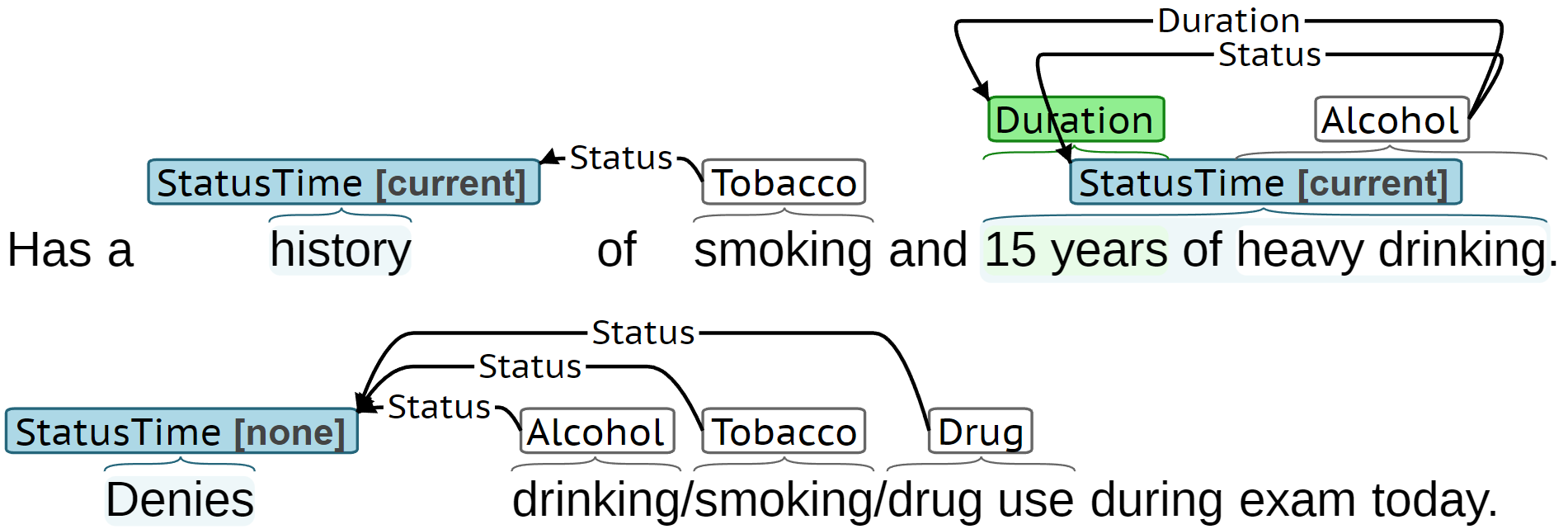}         
               \end{minipage}
           }
           \caption{Multiple gold \textit{Tobacco} and \textit{Alcohol} events in note with conflicting \textit{Status Time} labels.}
           \label{error_analysis_examples_b}
       \end{subfigure}

       \begin{subfigure}[b]{6in}
           \framebox{
               \begin{minipage}[t]{5.75in}
                   \includegraphics[scale=0.30]{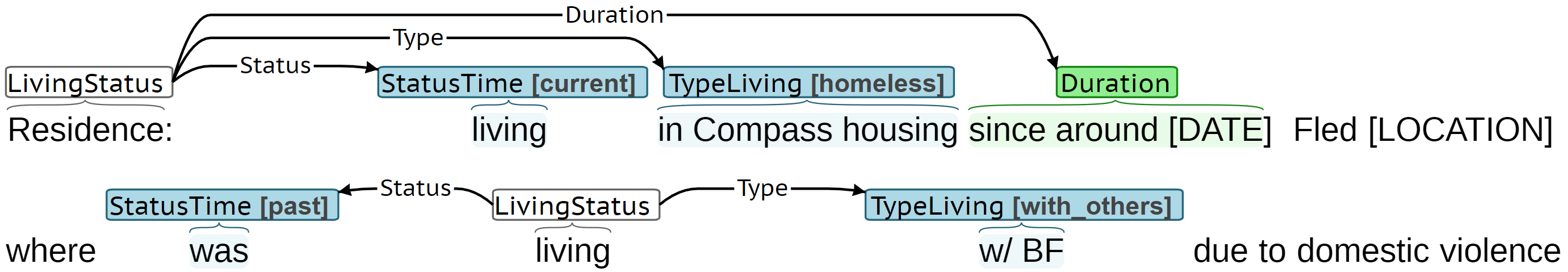}
               \end{minipage}
           }
           \caption{Multiple gold \textit{Living Status} events in a sentence with conflicting \textit{Status Time} and \textit{Type Living} labels.}
           \label{error_analysis_examples_c}
       \end{subfigure}

   \caption{Error analysis examples from Subtask C ($\mathcal{D}_{test}^{uw}$).}
   \label{error_analysis_examples}
\end{figure}

Formatting, linguistic, and content differences between the MIMIC and UW data resulted in lower performance in Subtask B (train on MIMIC, evaluate on UW), relative to Subtask A (train and evaluate on MIMIC). The inclusion of UW data improved performance in Subtask C (train on MIMIC and UW, evaluate on UW), largely mitigating the performance degradation. We manually reviewed the predictions from the top three teams in each subtask (same data as Figures 3-5
) focusing on the notes with lower performance, to explore domain mismatch errors and challenging classification targets. 

The UW data include templated substance use, which differs in format from MIMIC. In Subtask B, unpopulated templates, like `` Tobacco use: \_ Alcohol: \_ Drug Use: \_,'' resulted in false positives. There are differences in the substance use vocabulary that resulted in false negatives in Subtask B. 
For example, the UW data include descriptions of ``toxic habits’’ and uses ``nicotine’’ as a synonym for tobacco. 

The UW data contain living situation descriptions that lack sufficient information to resolve the \textit{Type Living} label, a necessary condition for a \textit{Living Status} event. In Subtask B, there were frequent false positives associated with living situation descriptions, like ``Lives in a private residence,’’ which are not present in MIMIC. The MIMIC data include homogeneous descriptions of housing instability (e.g. ``homeless’’), whereas the UW data include more heterogeneous descriptions (e.g. ``undomiciled’’ or ``living on street’’) and references to regional named shelters (e.g. ```Union Gospel Mission’’). In Subtask B, this difference resulted in false negatives and misclassified \textit{Type Living}. The UW data also include residences that are uncommon in the MIMIC data, like skilled nursing facilities or adult family homes, contributing to false negatives in Subtask B. 

\textit{Employment} performance in Subtask B was negatively impacted by references to the employment of family members, confusability between having a disability and receiving disability benefits, and linguistically diversity. The linguistic diversity was especially challenging when a common trigger phase, like ``works’’, is omitted, as in ``Cleans carpets for a living.’’

More nuanced SDOH descriptions, where determinants are represented through multiple events, were more challenging classification targets. Figure 6 
presents gold  examples from Subtask C that were challenging targets. Figure 6a 
presents a sentence describing drug use, including the denial of use and positive toxicology. Extraction requires creating separate events for the drug denial and positive toxicology report, including correctly identifying the appropriate triggers among several commonly annotated trigger spans (``Drug Use,’’ ``drug use,’’ and ``polysubstance use’’). Figure 6b 
describes smoking and alcohol use history in the first sentence and refutes substance use in the second sentence. SHAC annotators annotated events using intra-sentence information, where possible, resulting in the conflicting \textit{Status Time} labels. Extraction requires identifying multiple triggers and resolving the inconsistent \textit{Status Time} labels. Figure 6c 
presents a \textit{Living Status} example where separate events are needed to capture previous and current living situations and there are multiple commonly annotated trigger spans (``Residence’’ and ``living’’). Here, homelessness must be inferred from ``Compass housing,’’ a regional shelter. The nuanced descriptions of SDOH in Figure 6 
are also more challenging to annotate consistently, contributing to the extraction challenge.

\ifsubfile
\bibliography{mybib}
\fi

\section*{Discussion}

The top-performing submissions utilized pretrained LM. A majority of the submissions, including the top-performers, utilized a combination of sequence tagging and text classification. However, the best performing system across subtasks was Microsoft’s seq2seq approach. The success of Microsoft’s seq2seq approach may be related to the classifier architecture (seq2seq vs. sequence tagging and text classification), LM architecture (T5 vs. BERT), data augmentation (additional shorter training samples), LM pretraining, and/or use of additional supervised data. Systems that utilized data-driven neural approaches performed better than systems based on rules, knowledge sources, and n-grams. 

The top-performing teams demonstrated good domain generalizability to the UW data in Subtask B ($\mathcal{D}_{dev}^{uw}$) for alcohol, drug, tobacco, and employment, with lower generalizability for living status. Incorporating in-domain data in Subtask C increased performance, resulting in similar performance in both domains (MIMIC for Subtask A and UW for Subtask C). The error analysis indicates SDOH risk factors tend to be extracted with lower performance than protective factors: i) performance for current and past substance use is lower than substance abstinence, ii) performance for current and past drug use is lower than for alcohol and tobacco, iii) performance for being on disability is lower than other employment categories, and iv) performance for homelessness is lower than living with family or (stably) alone. The error analysis also indicates extraction performance decreases as the density of SDOH events increases. Qualitatively, notes with more SDOH events tend to express more severe social needs (e.g. polysubstance use, long history of tobacco use, and more frequent living situation transitions). The analysis of SDOH factors (Figure 4
) and event density (Figure 5
) suggests notes describing higher risk SDOH are more challenging extraction targets.

\ifsubfile
\bibliography{mybib}
\fi

\section*{Conclusions}

SDOH impact individual and public health and contribute to morbidity and mortality. The \challengeName{} explores the automatic extraction of substance use, employment, and living status from clinical text using SHAC.\cite{LYBARGER2021103631} The top-performing teams in this extraction challenge used pretrained LM, with most using a combination of sequence tagging and text classification. A novel seq2seq approach achieved the best performance across all subtasks at 0.901 F1 for Subtask A, 0.774 F1 Subtask B, and 0.889 F1 for Subtask C. SDOH risk factors tend to be more challenging extraction targets than protective factors. Future work related to SDOH extraction should consider this potential for bias, especially for patients with higher-risk SDOH.

\ifsubfile
\bibliography{mybib}
\fi

\section*{Acknowledgments}
This work was done in collaboration with the UW Medicine Department of Information Technology Services.

\section*{Funding Statement}
This work was supported in part by the National Institutes of Health (NIH) National Center for Advancing Translational Sciences (NCATS) (Institute of Translational Health Sciences, Grant Nr. UL1 TR002319), the NIH National Library of Medicine (NLM) Grant Nr. R13LM013127, the NIH NLM Biomedical and Health Informatics Training Program at UW (Grant Nr. T15LM007442), and the Seattle Flu Study through the Brotman Baty Institute. The content is solely the responsibility of the authors and does not necessarily represent the official views of the NIH. 


\section*{Competing Interests Statement}
The authors have no competing interests to declare.

\section*{Contributorship Statement}
All authors contributed to the organization of the shared task and the n2c2 workshop where the shared task results were presented. KL performed the data analysis and drafted the initial manuscript. All authors contributed to the interpretation of the data, manuscript revisions, and intellectual value to the manuscript.

\section*{Data Availability Statement}
The SHAC data set used in the shared task will be made available through the University of Washington.


\bibliography{mybib}

\pagebreak

\newpage
\section*{Appendix}

\subsection*{Extended Error Analysis}

Figures 7-9 
present the performance  by the number of gold events per note for a given event type (referred to as \textit{event density}): one (1), two (2), and three or more (3+) events per note. These figures also includes the total number of gold events ({\small \ding{58}}).

\begin{figure}[!ht]
    \centering
    \frame{\includegraphics[width=6in]{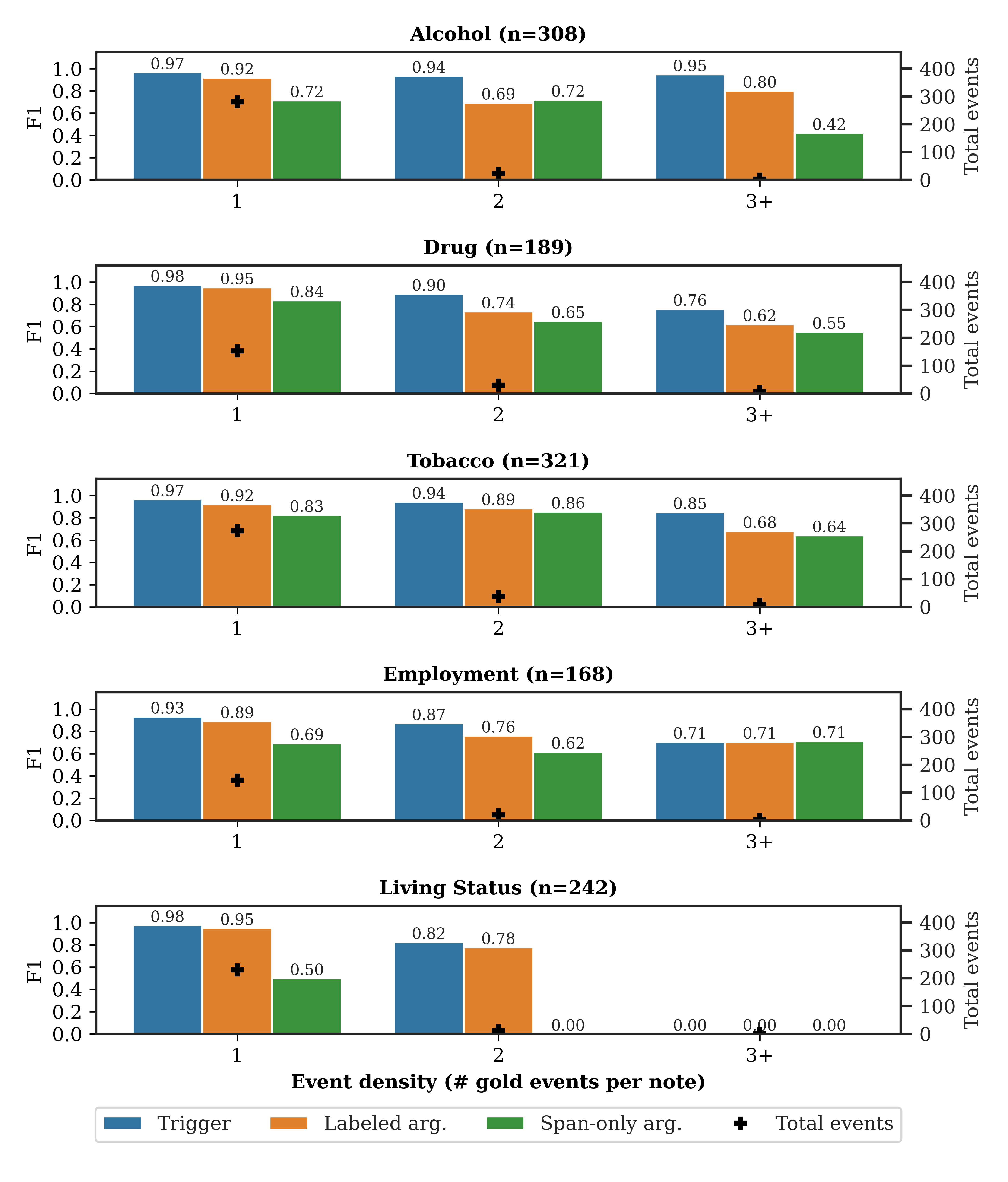}}
    \caption{Performance breakdown by event density for Subtask A ($\mathcal{D}_{test}^{mimic}$). The left-hand y-axis is the micro-averaged F1 for triggers, labeled arguments, and span-only arguments (vertical bars). The right-hand y-axis is the total number of gold events (\ding{58}).}
    \label{density_a}
\end{figure}

\begin{figure}[!ht]
    \centering
    \frame{\includegraphics[width=6in]{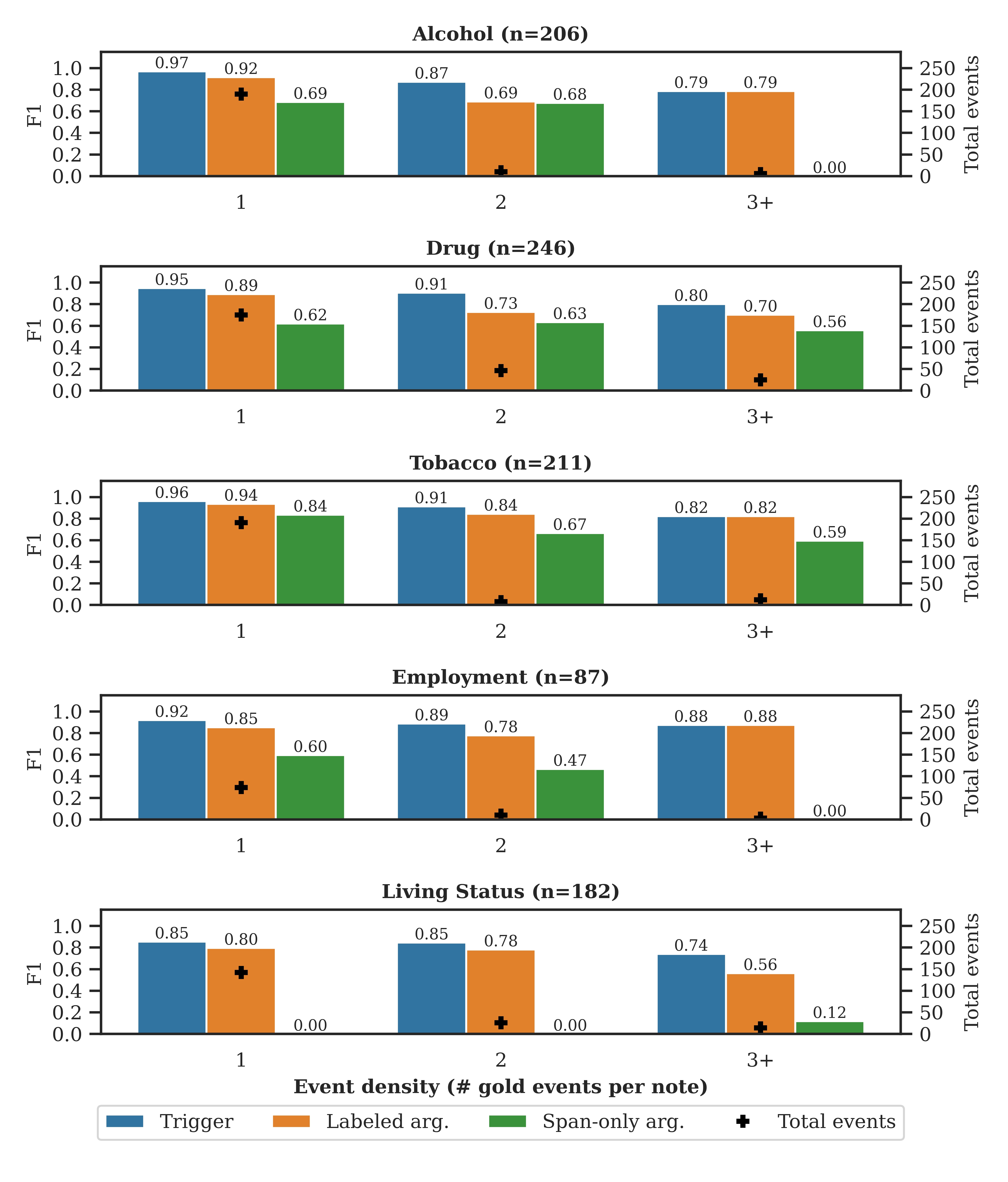}}
    \caption{Performance breakdown by event density for Subtask B ($\mathcal{D}_{dev}^{uw}$). The left-hand y-axis is the micro-averaged F1 for triggers, labeled arguments, and span-only arguments (vertical bars). The right-hand y-axis is the total number of gold events (\ding{58}).}
    \label{density_b_dev}
\end{figure}

\begin{figure}[!ht]
    \centering
    \frame{\includegraphics[width=6in]{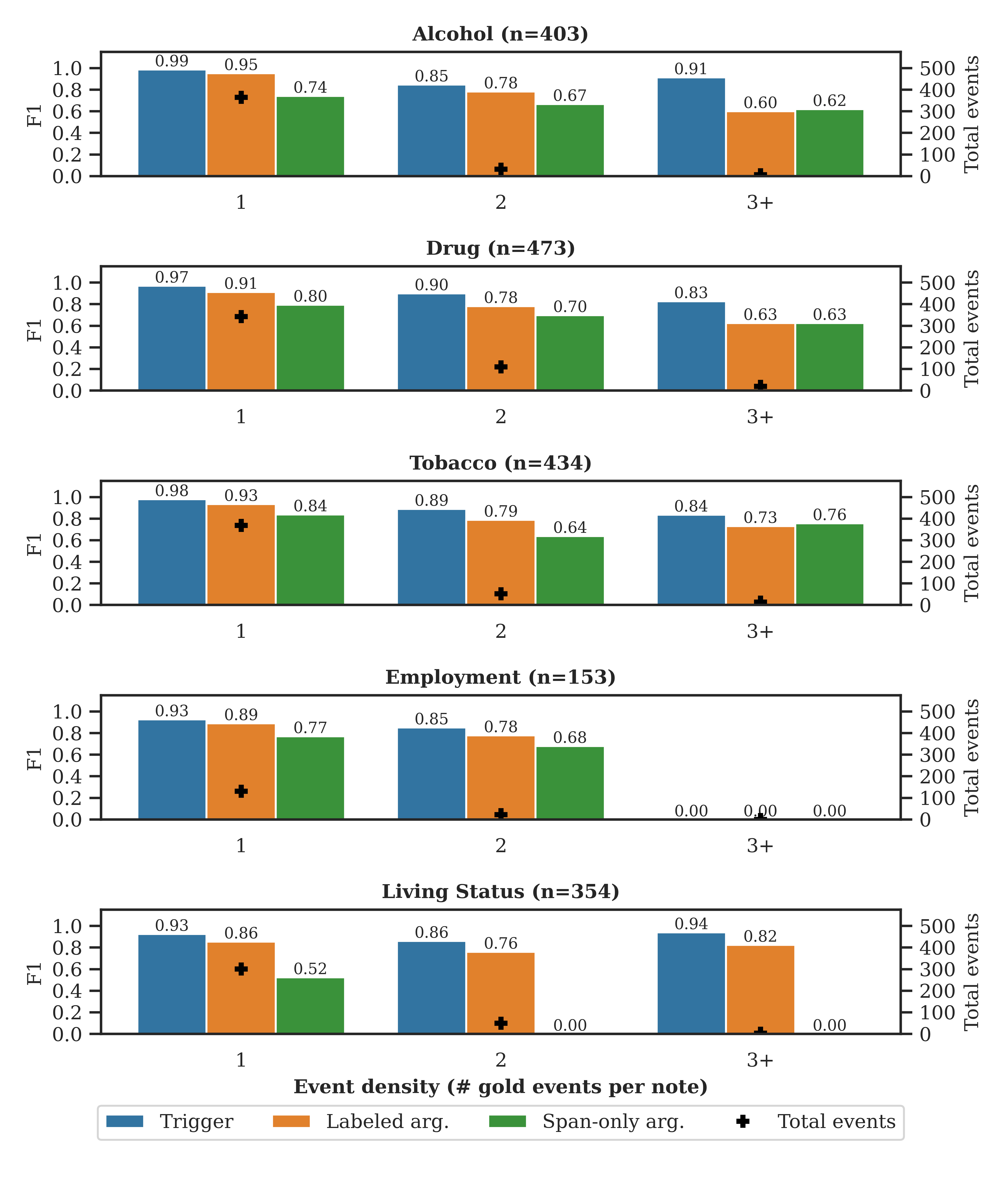}}
    \caption{Performance breakdown by event density for Subtask C ($\mathcal{D}_{test}^{uw}$). The left-hand y-axis is the micro-averaged F1 for triggers, labeled arguments, and span-only arguments (vertical bars). The right-hand y-axis is the total number of gold events (\ding{58}).}
    \label{density_c}
\end{figure}

\ifsubfile
\bibliography{mybib}
\fi


\end{document}